\documentclass[journal,article,accept,moreauthors,pdftex]{Definitions/mdpi} 
\firstpage{1} 
\pubvolume{1}
\issuenum{1}
\articlenumber{0}
\pubyear{2021}
\copyrightyear{2020}
\datereceived{} 
\dateaccepted{} 
\datepublished{} 

\Title{AI-enabled Efficient and Safe Food Supply Chain}

\TitleCitation{AI-enabled Safe and Efficient Food Supply Chain}

\Author{Ilianna Kollia $^{2, \ddagger}$\orcidA{}, Jack Stevenson $^{1}$\orcidB{},  Stefanos Kollias $^{1, 2}$\orcidD{}}

\AuthorNames{Ilianna Kollia, Jack Stevenson and Stefanos Kollias}

\AuthorCitation{Kollia, I.; Stevenson J.; Kollias, S.}

\address{%
$^{1}$ \quad School of Computer Science, University of Lincoln, Lincoln, UK\\
$^{2}$ \quad School of Electrical and Computer Engineering, National Technical University of Athens, Athens, Greece
}
\corres{Correspondence: ilianna2@mail.ntua.gr} 

\abstract{This paper provides a review of an emerging field in the food processing sector, referring to  efficient and safe food supply chains, 'from farm to fork', as enabled by Artificial Intelligence (AI). Recent advances in machine and deep learning are used for effective food production, energy management and food labeling.  Appropriate deep neural architectures are adopted and used for this purpose, including Fully Convolutional Networks, Long Short-Term Memories and Recurrent Neural Networks, Auto-Encoders and Attention mechanisms, Latent Variable extraction and clustering, as well as Domain Adaptation. Three experimental studies are presented, illustrating the ability of these AI methodologies to produce state-of-the-art performance in the whole food supply chain. In particular, these concern: (i) predicting plant growth and tomato yield in greenhouses, thus matching food production to market needs and reducing food waste or food unavailability; (ii) optimizing energy consumption across large networks of food retail refrigeration systems, through optimal selection of systems that can get shut-down and through prediction of the respective food de-freezing times, during peaks of power demand load;  (iii) optical recognition and verification of food consumption expiry date in automatic inspection of retail packaged food, thus ensuring safety of food and people’s health.}


\begin{document}
\section{Introduction}
Food and drink processing is one of the largest manufacturing sectors worldwide, including all processing steps 'from farm to fork' \cite{ref105}. The economic values of the AI-enabled precision farming market is estimated to grow and reach EUR 11,8 billion by 2025 globally \cite{ref104}.
 There are significant challenges within these processing steps, regarding waste, safety and energy use \cite{ref62}. Artificial Intelligence (AI) and Machine Learning (ML) technologies offer a transformative solution and recent Deep Learning (DL) approaches have innovated AI-enabled efficient yield and food production, food conservation and supply, reducing food waste and improving food safety. 

This paper presents  recent progress in the food production and supply pipeline
achieved through the use of AI and DL technologies, focusing on three main tasks of the 'from farm to fork' pipeline:  a) accurate prediction of yield growth and production in greenhouses; b) optimization of power consumption in food retailing refrigeration systems; c) quality control in retail food packaging.

At first, accurately predicting yield and food production is of great significance for reducing food waste and achieving smooth supply of food to supermarkets.  Crop growers nowadays prefer using greenhouses than field growing. By using greenhouses, growers can extend the growing
season and protect crops against changes of weather and  temperature.  In addition, in greenhouses, environmental parameters, such as temperature,  humidity, radiation, carbon dioxide, soil quality and fertilization, can be controlled, providing a safe environment for crop growing \cite{ref74, ref108}.

Developing models which
can effectively predict growth and yield can help
growers improve the environmental control for
better production, match supply and market
demand and lower costs \cite{ref70, ref71}.
Greenhouse farming operations, yield and production prediction still rely heavily on human expertise. Automated yield and production prediction systems can let growers effectively anticipate weekly fluctuations and avoid problems of both over-demand and overproduction arising if the yield cannot be accurately predicted \cite{ref57, ref58}.

As with many bio-systems, plant growth is a
highly complex and dynamic environmentally
linked system. Therefore, growth and yield
modeling is a significant scientific challenge \cite{ref107, ref72}.
Modeling approaches vary in a number of aspects, including, scale of interest, level of description and integration of environmental stress. There is a large number
of tools that can help farmers in making decisions. These can provide yield rate prediction,
suggest climate control strategies, or synchronise
crop production with market demands. 

According to \cite{ref73, ref56} two basic modeling approaches are possible, namely, "knowledge-driven" and "data-driven" modeling. The knowledge driven approach relies mainly on existing domain knowledge, including biophysical models for specific species and plants \cite{ref29, ref30}. In contrast, a data-driven modeling approach is capable of formulating a model solely from gathered data without necessarily using domain knowledge \cite{ref60}.
In this paper we focus on recently developed deep
learning models, which can be  trained with environmental data ($CO_2$, humidity,
radiation, outside temperature, inside temperature), as well as data for the actual yield and for significant plant characteristics, so as  to accurately predict the yield, or these characteristics. 

In the following phase of the pipeline, food enters the retailing phase, during which is stored in refrigeration systems of supermarkets. Food refrigeration accounts for a large percentage of electricity demands in developed countries, e.g., it is over 14\% of UK’s electricity demands, and mass refrigeration is responsible for a large percentage of carbon emissions, e.g. around 12\% in UK. When energy consumption exceeds certain limits, this can cause unmanageable spikes in energy usage, with unpredictable problems \cite{ref82, ref84}. To reduce load, countries impose response times on large retail energy consumers, requiring them to react urgently or face high financial penalties. For large companies with thousands of refrigerators, meeting this requirement relies on staff reactiveness, a highly manualized, resource intensive, non-synchronous procedure . 

A procedure to cope with such cases is to turn off some of the refrigerators in such cases for some period of time \cite{ref90, ref91}. Using AI and machine learning, it is possible to predict which refrigerators to select and for how long to turn them off, whilst  maintaining food quality and safety. In this paper we focus on a deep learning approach that has been recently developed for coping with such situations.   

In the third phase of the pipeline, after assuring a safe and efficient storage of food in retailing refrigeration systems, the food is packaged and delivered to the shelves of  supermarkets, so that customers can buy and consume it. Serious problems are also met in this phase as well, due to large amounts of food that remain unsold beyond their expiration date. Up to 30\% of food is wasted each year, and food poisoning (including over 64,000 annual incidents in the UK from Campylobacter alone) is costly for the national healthcare system and individuals affected \cite{ref75}. 

Food manufacturing faces the risk of product recalls and emergency product withdrawals  caused by human error on packaging lines; if the expiry date is incorrectly listed as too early, consumers may believe that the product has reached the end of its shelf life and not consume, i.e., waste it. Conversely, if the expiry date exceeds the actual date, consumers may use the product beyond its safe timeframe, risking illness or potential fatality. 

Due to prevalence of inkjet printers in food industry, characterized by high degree of quality variability, traditional Optical-Character-Recognition based vision systems have not been widely implemented, as they struggle with varied or distorted text. In this paper we focus on recently developed deep learning systems that can identify and verify the presence and legibility of expiry date on food packaging photos captured while the products pass along production lines \cite{ref109, ref110}.

Section 2 presents the methods developed and used to implement the above described frameworks. Section 3 describes the application of the developed deep learning methodologies to real life environments and the achieved performance in all three phases of the pipeline. Discussion of the results that have been obtained and description of the future directions are provided in Section 4.  
\section{Materials and Methods}

\subsection{Fully Convolutional Networks} 

Fully Convolutional Networks (FCN) \cite{ref80} do not contain dense layers, like typical Convolutional Neural Networks (CNN), but contain 1x1 convolutions that perform the task of fully connected layers. They are used in the experimental Section for text detection in images. Their architecture includes three parts: a feature
extractor stem part, a feature merging branch part and the output
part.

The feature extractor stem part is  
a PVANet \cite{ref102}, including convolutional
and pooling layers. In our implementation, four feature map are extracted from an input image through the convolutional layers, enabling multi-scale detection
of text regions of different sizes. Moreover, each pooling layer down-samples a corresponding feature
maps by a factor of 2.

The extracted four feature maps, $f_{i}, i=1,\ldots,4$,
are then fed in the feature-merging branch part.  
In the $i$-th merging stage, the feature map
$f_{i-1}$ obtained in the former stage is fed to an unpooling
layer for size doubling and then concatenated
with the feature map $f_{i}$ into a new feature map. A
convolutional 1×1 operator is then applied, followed by a convolutional 3×3 operator that produces the merging stage output. 
Finally, a convolutional 3×3 layer produces the merging output and feeds it to the
output layer.
Multiple convolutional 1×1 operations are then used to produce the network outputs.

\subsection{Long Short-Term Memories}

Long Short-Term Memories (LSTM) are a variation of the Recurrent Neural
Network (RNN) architecture \cite{ref15}. Networks composed of LSTM units have been able to solve the gradient
vanishing problem met in long-term time series analysis. 

To achieve this, the LSTM structure
contains three modules: the forget gate, the input gate and the output gate.
The forget and input gates control which part of the information should be removed, or
reserved to the network; the output gate uses the processed information
to generate the provided output. LSTM units also include a Cell State, which allows
the information to be saved for a long time. 

\subsection{Convolutional-Recurrent Neural Networks}

Convolutional-Recurrent Neural Networks (CRNN) are lightweighted networks, which are used in the experimental section
for text recognition. They include three
parts, i.e., a feature extraction part, a  bidirectional
LSTM-RNN part and a transcription layer part.

The feature extraction part consists of a VGG network \cite{ref103}. An input image is divided into $T$ different image
patches; feature vectors $x_{1}, x_{2}, . . . , x_{T}$ are extracted from  
the different patches through convolutional and pooling
layers. These feature vectors
are then fed to a deep
bidirectional Recurrent Neural Network (RNN) composed of recurrent layers with LSTM units. The RNN part can
 captures contextual
dependencies between text in consecutive
image patches. It is also able to operate on arbitrary lengths of text
sequences.

The final layer in the CRNN is used for transcription, 
converting the text predictions made by the  bidirectional
LSTM-RNN into a label sequence. This is achieved through maximization of  a
conditional probability given the bidirectional LSTM-RNN
predictions  \cite{ref81}.

\subsection{Encoder-Decoder Model}

\begin{figure}[tph!]
\includegraphics[totalheight=7cm]{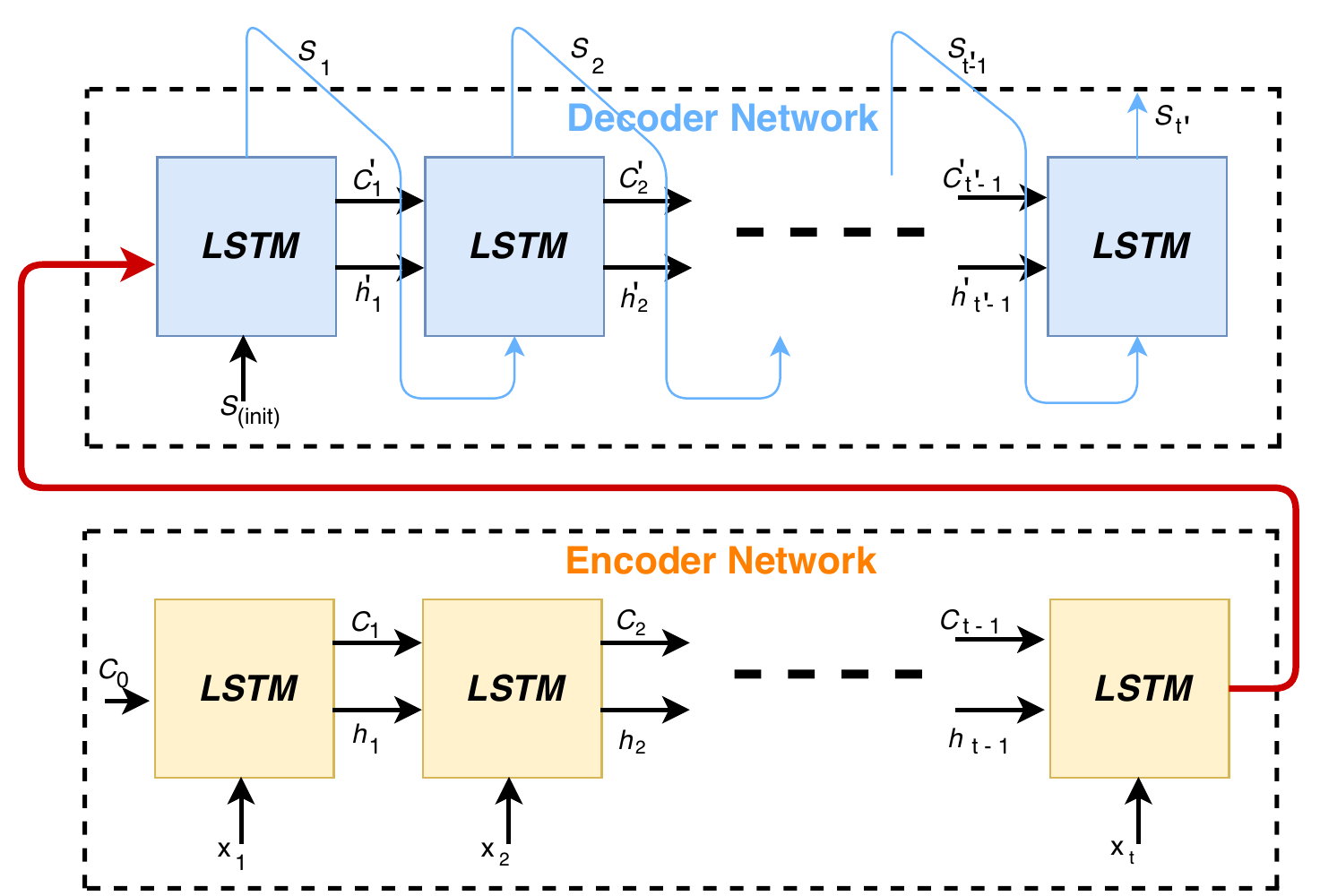}
\centering
\caption{LSTM encoder decoder architecture}
\label{fig:LSTM_Decoder_Encoder}
\end{figure}

In LSTM-based encoder-decoder models, the encoder part compresses the information
from the entire input sequence into a vector composed of the sequence
of the LSTM hidden states. Consequently, the encoder summarizes the
whole input sequence ($x_1, \ldots, x_t$) into the cell ($C_0, \ldots, C_{t-1}$) and memory ($h_1, \ldots, h_{t-1}$) state vectors and passes them to the decoder \cite{ref16}. The latter uses this representation as initial state to reconstruct the time series ($S_{t'}$).
 The architecture employs two LSTM networks called the encoder and decoder, as shown in  Figure \ref{fig:LSTM_Decoder_Encoder}.

.

\subsection{Attention Mechanisms}

Attention mechanisms help to focus on feature segments of high
significance \cite{ref100}. An attention mechanism \cite{ref101} models long-term dependencies by computing context
vectors as weighted sums of all provided information. Such dependencies can be computed across the different internal LSTM layers, as well as over
the LSTM output layers.

 Output Predictions  can  be  derived  using  the  conditional probability distribution of the input signal and of the previous samples of the output. These are computed in terms of  the current context, i.e., a vector holding information of which inputs are important at the current step. The context is derived from both the current state and the input sequence. 
 
\subsection{Performance Visualization}

Two methodologies are examined in the paper
for visualization of the performance of the developed prediction models, based either on class activation maps, or on latent variable extraction from trained deep models and adaptive clustering.

\subsubsection{Class Activation Mapping}

Deep neural networks are usually  viewed as black boxes which do not provide means to visualize, or explain the decision making process.   
Class activation mapping (CAM) \cite{ref17} and Gradient-weighted CAM \cite{ref47} are efficient ways to visualize the significance of  various parts of an input image for network training and decision making.

CAM-based methods provide such visualization by generating heat maps for each input image, focusing on the areas that influenced the network's prediction. In practice, they highlight the image pixels which are mainly used for classifying each image to a specific category.

\subsubsection{Latent Variable Adaptive Clustering} 

Extraction of latent variables from trained deep neural networks and generation of concise representations through clustering them has been recently used as another means for visualising and explaining the decision making process \cite{ref1, ref2, ref41}.  

Let us assume that for each  input $k$, we extract $M$ neuron outputs, as latent variables, from the trained deep neural network, forming a vector ${\textbf{v}}(k)$. In total, we get:

\begin{equation}
\label{eq:traininglatent}
\mathcal{V} = \big\{({\textbf{v}}(k), \ k=1,\ldots,N\big\} 
\end{equation}
where $N$ is the number of available training data. 

We generate a concise representation of the $\textbf v$ vectors, that can be used as a backward model, to trace the most representative  inputs for the performed prediction.  This can be achieved using a clustering algorithm, e.g., k-means++ \cite {arthur2006kICCV}, which generates, 
say, $L$ clusters  ${Q} =\{\textbf{q}_1,\ldots,\textbf{q}_L\}$ by minimizing the following criterion: 
\begin{equation}
\label{eq:kmeans}
\widehat{{Q}}_{k\text{-means}} = \underset{{Q}}{\operatorname{arg\ min}} 
\sum_{i=1}^{L} \sum_{\mathbf{v}\in {V}}^{} 
\big|\big|\textbf{v}-\textbf{$\mu$}_{i}\big|\big|^{2}
\end{equation}
where $\textbf{$\mu$}_{i}$ is the mean of $\textbf v$ values included in cluster $i$. 

Then, we compute each cluster center $\textbf{c}(i)$, generating the set $C$, which constitutes the targeted concise representation:

\begin{equation}
\label{eq: cluster centroid set}
\mathcal{C} = \big\{(\textbf{c}(i), \ i=1,\ldots,L\big\} 
\end{equation}

This representation can be used to provide analysis of variability between distributions and adapt information across different food datasets \cite{ref40}.

\subsection{Domain Adaptation}

Most of the deep learning models are developed through supervised learning over large annotated datasets. However, labeling large datasets
consumes a lot of time and is  cost ineffective. In addition,
when deploying a trained model to a real-life processing pipeline, the
assumption that both the source (training set) and the
target one (application-specific environment) are drawn from the same
distributions, may not be valid. In such cases, the deep learning model,
trained on the source domain will not generalize well on the
target domain. This is known as domain shift and training a model in the presence of domain shift is known as Domain Adaptation \cite{ref51}.

Many domain adaptation methods have been proposed in the last few years. Discrepancy, Adversarial and
Reconstruction based approaches are the three primary domain
adaptation approaches currently being applied to address
the distribution shift \cite{ref18}. Discrepancy-based approaches rely
on aligning the distributions in order to minimize the divergence
between them. The most commonly used discrepancy-based
methods are Maximum Mean Discrepancy (MMD) \cite{ref19},
Correlation Alignment (CORAL) \cite{ref20} and Kullback–Leibler
divergence \cite{ref21}. Adversarial-based approaches minimize the
distance between the source and the target distributions using
domain confusion, an adversarial method \cite{ref22, ref23} inspired by Generative
Adversarial Networks (GANs) \cite{ref93}. Reconstruction-based approaches create
a shared representation between the source and the target
domains whilst preserving the individual characteristics of
each domain \cite{ref24, ref48}.

In the experimental Section we focus on the use of domain adaptation for ensuring safety and waste reduction in the food supply chain. The domain adaptation model
aims at minimizing the feature discrepancy, for
learning domain-invariant representations, the class boundary
discrepancy, for minimizing the mismatch among different classifiers,
and classification loss, for improving source data classification and leading to improved generalization
on the target dataset \cite{ref25}.

The model jointly adapts
features combining MMD with CORAL metrics, in order to align the underlying first and second order statistics.
MMD defines the distance between the source and target distributions,
with their mean embeddings in the Reproducing Kernel Hilbert
Space. The CORAL Loss is also  used to minimize the distance between the second order
statistics (covariances) of the source and target features.

The total feature discrepancy loss is therefore given as: 

\begin{equation}
\label{eq: loss}
Loss_{FD} = Loss_{MMD} + Loss_{CORAL}
\end{equation}

In case we are considering domain adaptation from  more than one source domains, the respective classifiers are likely to misclassify
the target samples near the class boundary, since they are
trained using different source domains. In such cases we also minimize the class discrepancy loss
among classifiers, $Loss_{CD}$ \cite{ref25}, making their probabilistic outputs
similar. Finally, since the network is trained with labeled source data, the classification, cross-entropy, loss, $Loss_{CL}$ is additionally minimized during training.

\begin{equation}
\label{eq: loss1}
Loss_{TOTAL} = Loss_{FD} + Loss_{CD} + Loss_{CL}
\end{equation}

\section{Experimental Study}

\begin{figure*}[tph!]
\includegraphics[totalheight=7cm]{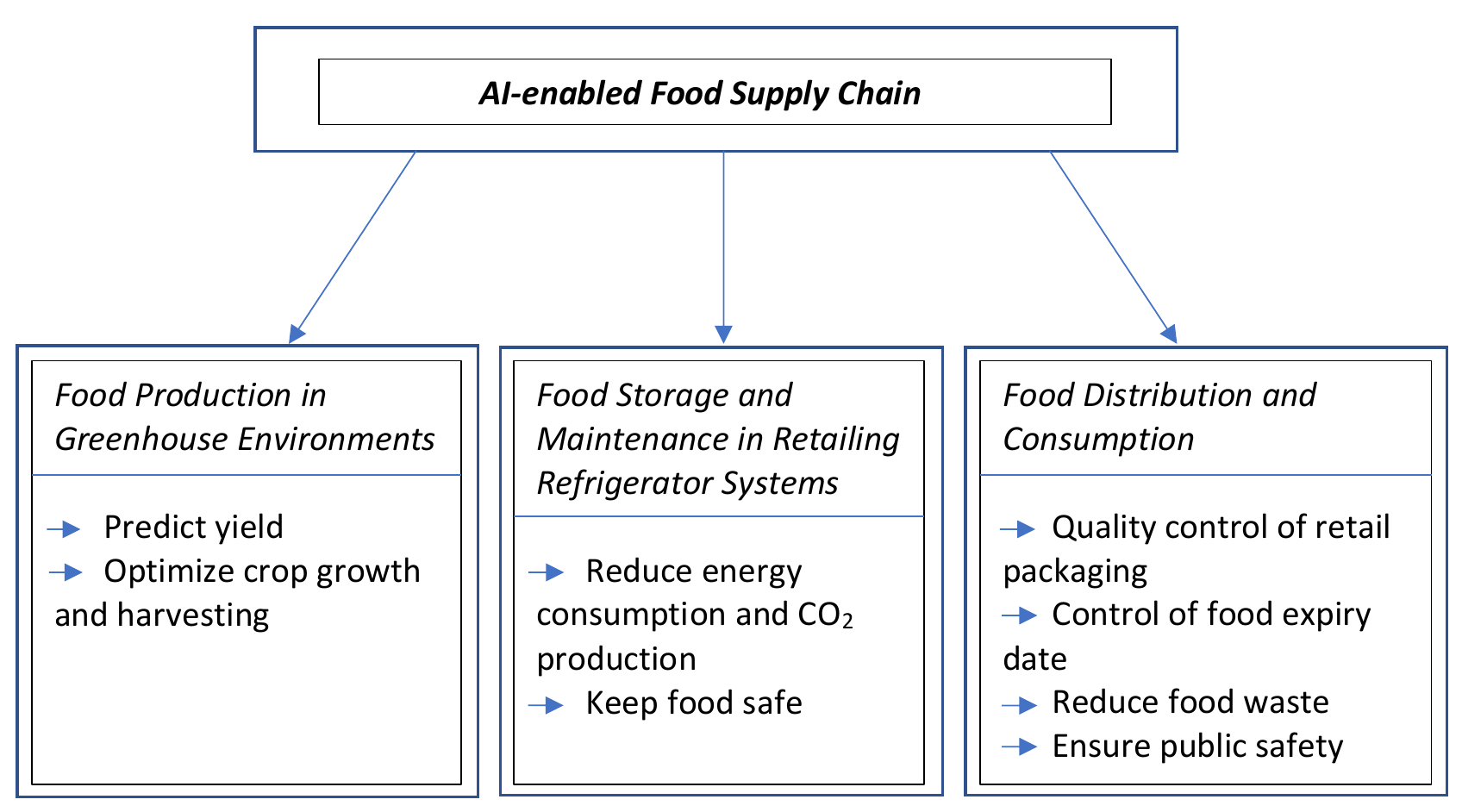}
\centering
\caption{Food Supply Chain}
\label{fig:food_supply diagram}
\end{figure*}

Figure \ref{fig:food_supply diagram} shows the described AI-enabled Food Supply Chain, including three pillars, which span food production, food storage and maintenance and  food distribution and consumption: 

- Food production in greenhouse environments, with a focus on predicting yield and optimizing crop growth and harvesting.

- Food storage and maintenance in retailing refrigerator systems, with a focus on reducing energy consumption and $CO_2$ production, whilst keeping food safe.

- Food distribution and consumption with a focus on quality control of retail packaging, through visual inspection of the food expiry date, while aiming to reduce food waste and avoid public health problems.

\subsection{Food Production in Greenhouse Environments}

\subsubsection{Plant Growth Prediction}
 
 Machine learning techniques have been used to predict growth of Ficus plants using  data  collected  from  four  cultivation  tables  in  a  greenhouse compartment of the Ornamental Plant Research Centre in Destelbergen, Belgium. Greenhouse microclimate was continuously  monitored, while controlling  the  window  openings, a thermal screen, an air heating system, assimilation light and a $CO_{2}$ adding system. In particular, ficus stem  diameter  was  continuously  monitored  on  three  plants and data were collected corresponding to its hourly  variation  rate, i.e.,  as  the difference between the current stem diameter and the stem diameter recorded on one  hour earlier.  The environmental data were also continuously recorded in hourly basis.

 The prediction problem concerns one-step, two-step and three-step forecasting of the stem diameter \cite{ref106, ref59}. In one-step forecasting, stem diameter measurements and environmental data collected, in a time window of the previous  15  hours, are used to predict the stem diameter value in the current hour. In two-step-ahead (6 hour) forecasting, the above data are used to predict the stem diameter six hours ahead. Three-step-ahead (12 hour) forecasting  makes a stem diameter prediction 12 hours ahead. 
 
\begin{figure*}[tph!]
\includegraphics[totalheight=11.5cm]{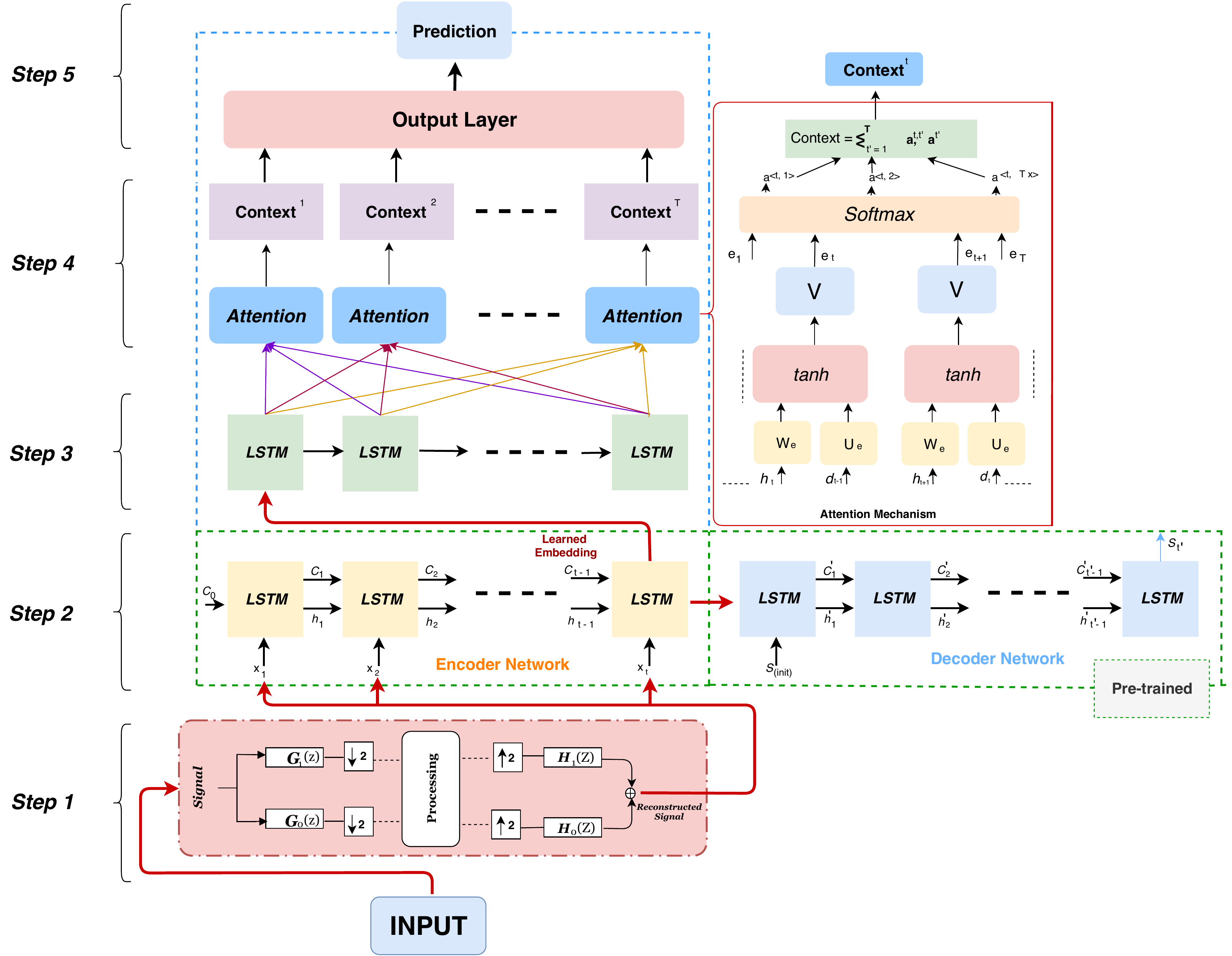}
\centering
\caption{Deep architecture  (WT-ED-LSTM-AM) for stem diameter prediction.}
\label{fig:stem_diagram}
\end{figure*}

Figure \ref{fig:stem_diagram} presents an effective stem diameter prediction method, based on combination of the  Encoder-Decoder (ED) model described in Section 2.4, the LSTM model described in Section 2.2. and the attention model (AM) described in Section 2.5 \cite{ref26}.

At first (Step 1), a denoising filter is applied to the input data, composed of the former stem diameter values and the respective environmental parameters. The filter is based on Wavelet Transform (WT) of the input data, removing the WT high frequency component, and generating a smoothed version of the input data \cite{ref120}. The ED model is then applied (Step 2). 
The encoder
is pre-trained to extract useful and representative embeddings from the
reconstructed time series data. A two-layer LSTM (128 and 32 neurons respectively) are used in the encoder shown in Figure 1.  The decoder 
learns to generate the (reconstructed) input signal from the embeddings, thus optimizing this feature extraction procedure. Then, the LSTM network, with 128 neurons (Step 3), with attention
mechanism (Step 4), is trained to model respective long-term dependencies, using the
learned embedding as input features. A single layer neural network (Step 5) provides the final one-step, or  multi-step ahead prediction provided by this WT-ED-LSTM-AM architecture.

 In the accomplished experiments, the first 70\% of data samples constituted the training set, the next 10\% of data samples were the validation set and the rest 20\% of data samples formed the test set. Min-max normalization was applied to the input data, scaling their values in the range [0, 1].

 The experimental results illustrate the very good performance of the described
method. Table \ref{table:ficus_Accuracy} shows that the method  outperforms all standard machine learning techniques, including  a Support Vector Regressor (SVR), a Random Forest Regressor (RFR), a two-layer LSTM and a Multilayer Perceptron (MLP) with Stochastic Gradient Descent; a learning rate ls = 0.001 and a batch size of 32 were adopted. All models were trained for 100 epochs, using the same training, as well as validation and test data sets. The Root Mean Squared Error (RMSE) was used as performance evaluation criterion. To illustrate the contribution of the WT and AM steps, Table \ref{table:ficus_Accuracy} also illustrates that the achieved performance gets worse, if either of these two steps is not included in the prediction approach.  

\begin{table}[t!]
\caption{Performance comparison of the WT-ED-LSTM-AM method to machine learning methods and ablation study}
\label{table:ficus_Accuracy}
\centering
\begin{tabular}{|c|c|c|c|}
\hline 
& \multicolumn{3}{c}{RMSE}\\
\hline
Method & One Step Prediction & Two Step Prediction & Three Step Prediction \\
\hline \hline
SVR & 0.65 & 0.70 & 0.82\\
RFR & 0.74 & 0.66 & 0.72\\
MLP & 0.0034 & 0.0045 & 0.0048\\
LSTM & 0.0031 & 0.0033 & 0.0054\\
WT-ED-LSTM & 0.0028 & 0.0033 & 0.0042\\
ED-LSTM-AM & 0.0034 & 0.0030 & 0.0046\\
WT-ED-LSTM-AM & 0.0026 & 0.0028 & 0.0029\\
\hline 
\end{tabular}
\end{table}


\begin{figure*}[tph!]
\includegraphics[totalheight=10cm]{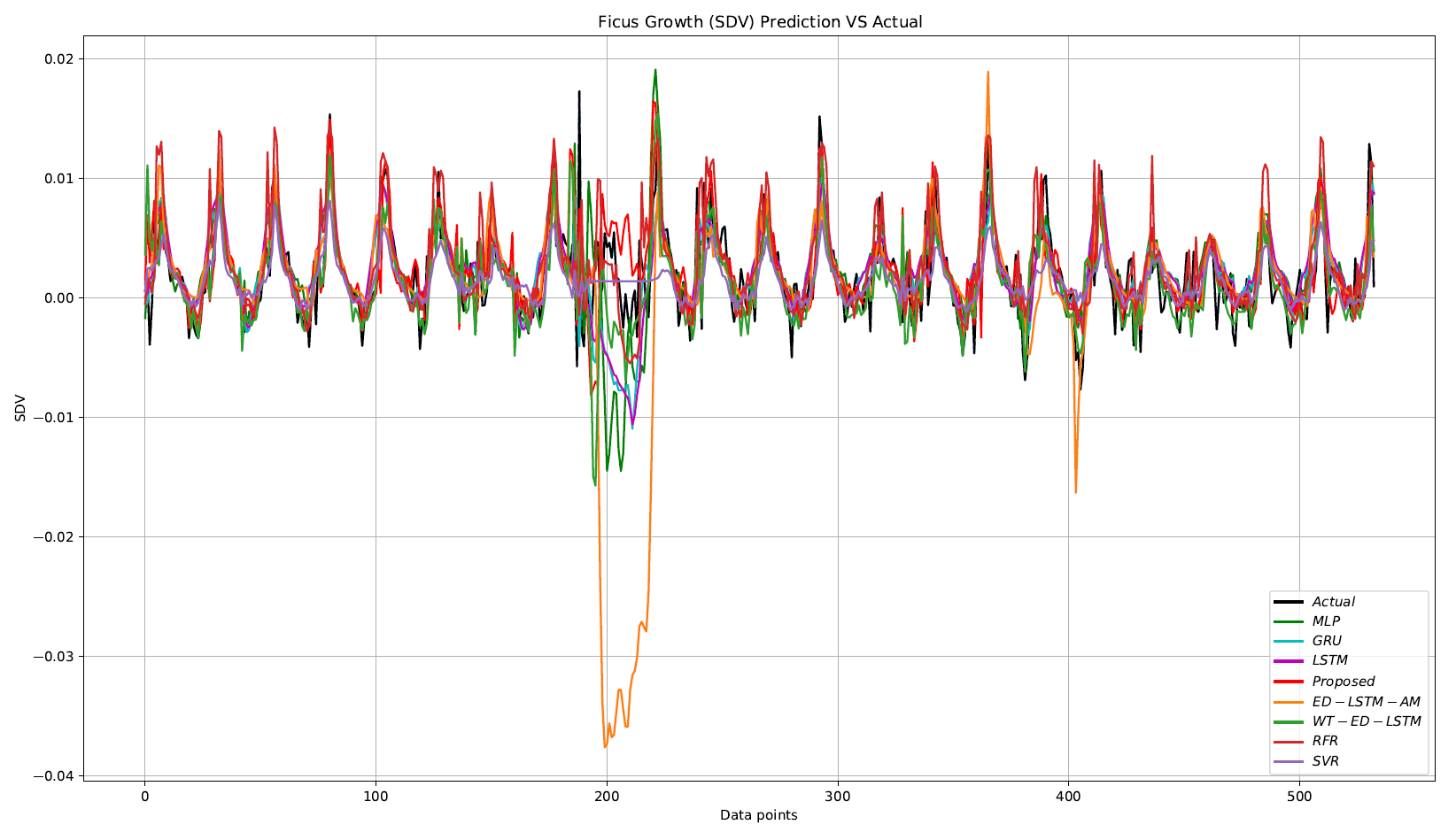}
\centering
\caption{Performance comparison in one-step prediction (described and standard ML methods).}
\label{fig5}
\end{figure*}

Figure \ref{fig5} shows the accuracy of Ficus growth one-step prediction by all methods
for about 600 data samples. It can be seen that the described model successfully
performs one-step ahead prediction, outperforming the other methods
and providing accurate estimates of almost all peak values in the original data.

\subsubsection{Yield Prediction}

Tomato crop growing in greenhouse environments is  a dynamic and complex system, with few models having been studied for it up to now \cite{ref27, ref28}. These models  are physics-dependent and represent biomass partitioning, crop growth, and yield as a function of several parameters. They are rather complex, with difficulty in estimating initial parameter values and  need calibration in every environment. 

\begin{figure*}[tph!]
\includegraphics[totalheight=8cm]{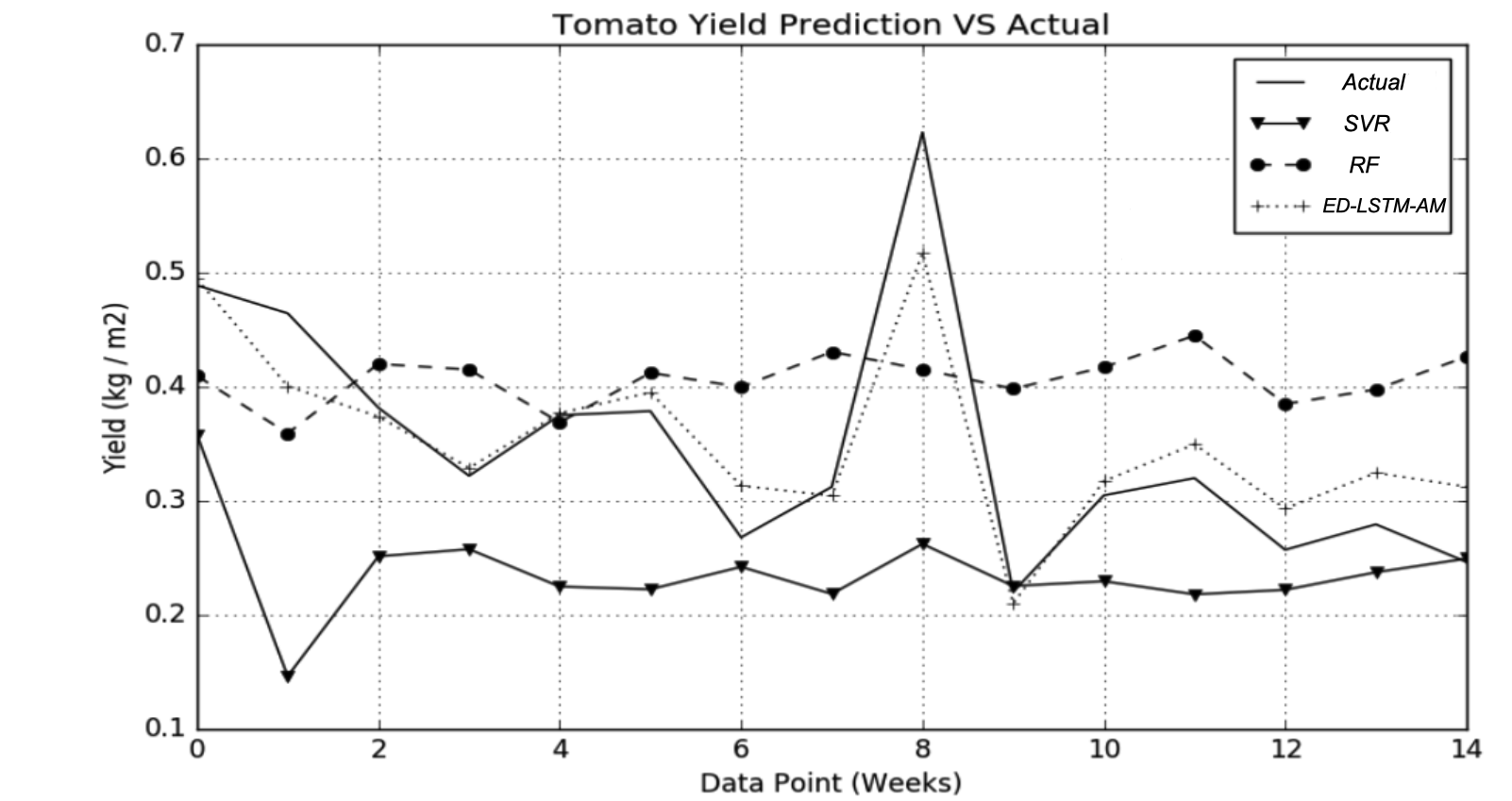}
\centering
\caption{Performance comparison in tomato yield prediction  (by ED-LSTM-AM  and standard ML methods).}
\label{figy}
\end{figure*}

The Tompousse model \cite{ref29} was developed to predict tomato yield in terms of the weight of harvested fruits, based on linear relationship between flowering rate and fruit growth. Another tomato yield model \cite{ref30} represented weekly yield fluctuations in terms of fruit size and harvest rate. Such models \cite{ref31} can help farmers in making  yield rate prediction, suggesting climate control strategies and synchronising crop production with market demands.

Machine learning models have been recently developed for tomato yield  prediction \cite{ref45}. Such models have been trained with data collected from Greenhouse farms in United Kingdom, over a period of two years, including both environmental parameters, i.e., $CO_{2}$, humidity, radiation, outside
temperature, inside temperature, as well as, yield actual measurements. 

The environmental data were collected on an hourly basis, while the yield on a weekly basis. To deal with these data characteristics,  data augmentation was performed, through interpolation, on the yield measurements, thus  resulting in daily data measurements. Averaging of the hourly environmental data was also performed, so as to achieve similar daily representations.

Experiments with these data also showed the ability of the above-described ED-LSTM-AM model (WT did not provide any significant improvement in this case) to provide accurate prediction of tomato yield. Figure \ref{figy} shows that this model outperforms standard ML methods, including SVR and RFR, in tomato yield prediction. The MSE in actual yield prediction has been 0.015 for SVR, 0.040 for RFR and only 0.02 for ED-LSTM-AM.

\subsection{Food Retailing Refrigeration Systems}

Currently the majority of deep learning frameworks  lack, or are still in their infancy, with regard to distribution. Novel solutions such as TensorFlow-distribute, or IBMs distributed deep learning are emerging to face this issue. The ability to share data between distributed nodes implementing, for example, the LSTM networks described in Section 2.2, is a critical bottleneck, despite the prevalence of databases providing very mature and advanced functionalities, such as MongoDB’s aggregate pipeline. 
There are few parallelisable deep learning frameworks, which can broadly be categorised as follows \cite{ref32}:

- Model parallelism  (TensorFlow, Microsoft CNTK), where a single deep model is trained using a group of hardware instances and a single data set.

- Data parallelism, where each hardware instance is trained across different data. 

- Hybrid parallelism, where a group of hardware trains a single model, but multiple groups can be trained simultaneously with independent data sets. 

- Automatic selection, where different parts of the training/test process are tiled, with different forms of parallelism between tiles.

In the food supply chain, machine learning methods can be effectively used in a hybrid parallel framework for controlling massive amounts of food retailing refrigeration systems. This was shown in a recent implementation \cite{ref33}, in which multiple LSTM models were coupled with a MongoDB database \cite{ref92} and were trained, using 110,000 real-life datasets - from about 1000 refrigerators.
 
 \begin{figure*}[tph!]
\includegraphics[totalheight=10cm]{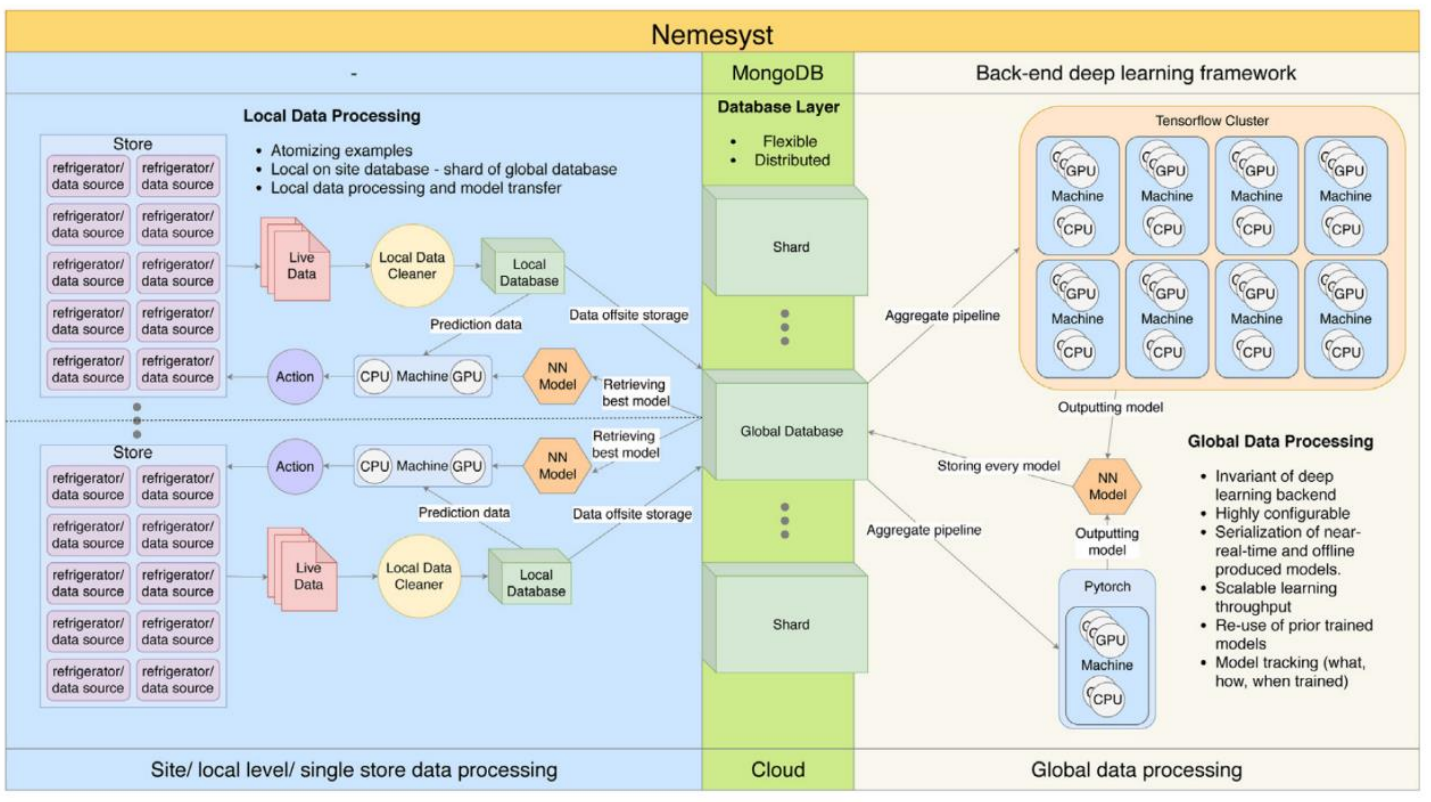}
\centering
\caption{The Nemesyst system for AI-enabled power supply control of retailing refrigerators}
\label{nemes}
\end{figure*}
 
 The generated Nemesyst system \cite{ref99} has been capable of predicting which refrigerators to select and how long to turn them off, whilst maintaining food quality and safety, in a Demand Side Response setting that modifies power demand load proportionally to available energy, in the National Grid of United Kingdom \cite{ref83, ref85}. The target of the research is to show how to optimise refrigeration systems with machine learning at scale, whilst ascertaining that food temperature does not pass certain  thresholds. 

The Nemesyst system simultaneously wrangles, learns, and infers multiple models, with multiple and distinct data sets, across a network of refrigerator systems. 
In such systems the thermal inertia/mass of food acts as a store of cold energy. However, the thermal inertia in a retail refrigeration case changes, if food is actively shopped by consumers and then refilled, if ambient temperature changes, if networks of stores possess multiple refrigeration systems, or if different types of food need specific control mechanisms \cite{ref94, ref95}. 

Deep learning models, such as LSTMs, can model thousands of assets simultaneously, whilst being retrained to select which refrigerators to shed electrical load/turnoff across a massive pool. This requires an algorithm that can predict the thermal inertia in individual cases and therefore how long they can be shutdown (typically up to a maximum of up to 30 min), ensuring food safety and contributing to decarbonisation.

As shown in Figure \ref{nemes}, a backend deep learning framework aggregates data from a distributed global database layer and trains with them multiple deep models, such as LSTMs. It then stores the trained models in the database layer. The latter interacts with each local store and its refrigerator systems, which share local database instances and retrieve the best deep network from the global database layer.

In the experimental study the target was to predict the time (in seconds) until the refrigerator temperature rises from the point it is switched off until it breaches a food safety threshold. This duration varies between cases, for example the threshold is heuristically $8^{o}C$ for high temperature refrigerators, and $-183^{o}C$ in low temperature refrigerators (freezers).  Since  the threshold for each fridge is not known, a two layer LSTM network is trained to predict the final point of defrost  from the data as this is the most consistent point we can rely on. 

As a consequence, a subset of fridges can be selected from the whole population, which are capable of remaining cool through the event requiring energy reduction. Knowing how long the refrigerators can remain off, before reaching this threshold, identifies the best candidates which reduce the power consumption sufficiently enough, while minimising the total number of affected refrigerators and potentially affected stock. The 110,000 defrost examples were split such that  10,000 were used for final testing, 10,000 were randomly selected each time for validation and the rest were used for training the system.

\begin{figure*}[tph!]
\includegraphics[totalheight=10cm]{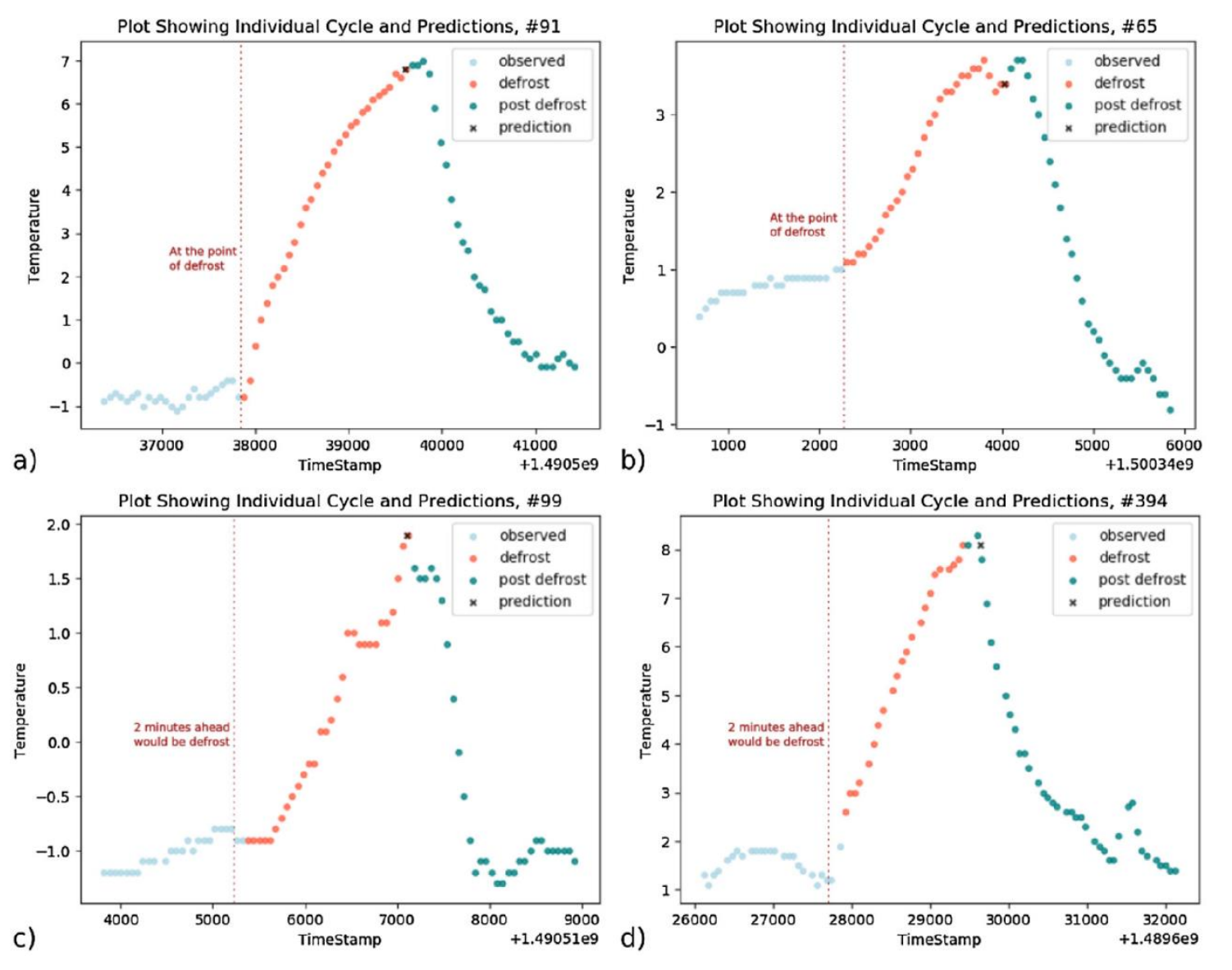}
\centering
\caption{Prediction of the defrost time in four refrigerators: “Observed” (light blue) denotes  training data;  “defrost” (orange) is the ground truth; “prediction” (×) is the prediction of the final ground truth (orange) value. }
\label{nemres}
\end{figure*}

Figure \ref{nemres} illustrates that an excellent prediction of the defrost time is achieved in four different cases.
The difference between the last “observed” and final “defrost” value is the time (in seconds) that defrost can occur for before reaching the unsafe threshold of $8^{o}C$. Images (a) and (b) show cases where prediction takes place at the point of defrost; images (c) and (d) show cases where prediction occurs 2 minutes in advance of defrost.

\subsection{Quality Control in Retail Food Packaging}

The requirement for food high quality and safety of food and the public 
is a very important issue during the whole food supply chain \cite{ref46}.
The food product information printed on the food package 
is very important for food safety. Incorrectly labelled product
information on food packages, such as the expiry date,
can cause food safety incidents, like food poisoning.  Moreover, such faults will incur high
reputation and financial cost to food manufacturers, also causing
large product recalls. As a consequence,  verification of the correctness of
the expiry date printed on food packages is of great significance .

In the following, it is shown that, instead of relying on operators to check the date code, an automated solution,  taking photos of each date code, can verify the programmed date code for that product run, allowing food processing businesses to introduce unmanned operations and achieve very high inspection with full traceability,  without compromising product safety. The production line will stop, if date code is incorrect, ensuring that the respective products  are not released into the supply chain, protecting consumers, business margins and their brand, while  reducing labour costs and food waste. 

The Food Packaging Image dataset used next consists of more than 30,000 images classified in two categories (existing valid date and non-existing or non-valid date) from six different locations in Midlands in United Kingdom. The training process was carried out on a 70 \% sample
with another 10\% used for the validation process. Finally, the remaining 20\% of the images formed the test set. Representative images are shown in Figure 8, showing  complete date (a), partial dates (b-c), unreadable date (d) and no existing date (e).

\begin{figure*}[tph!]
\includegraphics[totalheight=3cm]{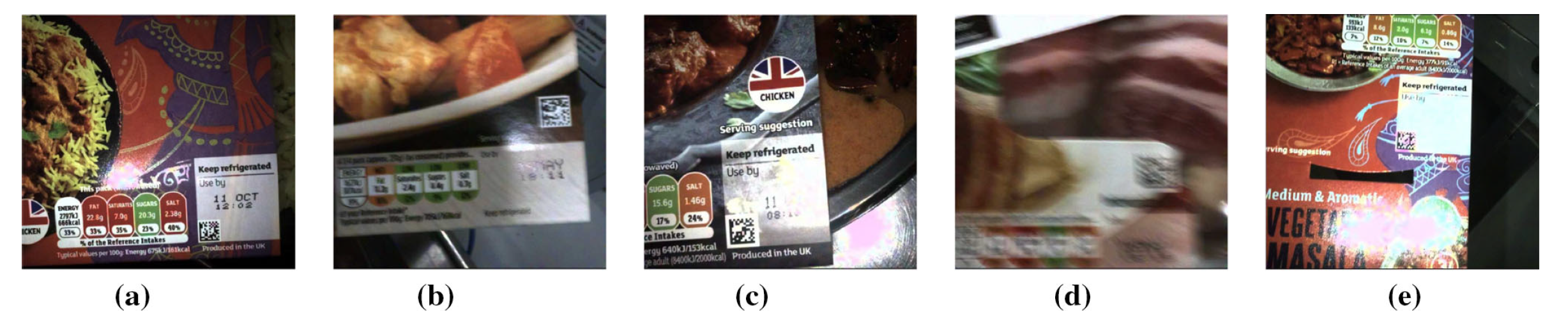}
\centering
\caption{Representative examples of food packaging images: (a) Complete date (day and month visible); (b) Partial date (no day visible); (c) Partial date (no month visible); (d) Unreadable; (e) No date.}
\label{dataset}
\end{figure*}

\subsubsection{The FCN-CRNN Approach for Expiry Date Recognition}

Optical character
recognition (OCR) systems \cite{ref34, ref49} can be applied to automatically
recognize expiry date characters based on food package images captured by RGB cameras. However, existing OCR systems can not perform effectively in
real-world expiry date recognition scenarios, with high variability, different fonts/angles, complicated 
designs with rich colours/textures, blurred
characters poor lighting conditions in food manufacturing/
retailer sites. Deep neural networks have been recently used as a means to tackle such problems \cite{ref76, ref77}.

In the following we present an approach composed of the FCN network, described in Section 2.1 and the CRNN network described in Section 2.3. Both networks are lightweighted and
achieve good performance for text detection and recognition
‘in-the-wild’, are fine-tuned and combined together to
detect and recognize the expiry date. Fine-tuning is performed through transfer learning  \cite{ref35}, by adapting a model pre-trained for
recognition of text, to recognition of expiry date.

\begin{figure*}[tph!]
\includegraphics[totalheight=10cm]{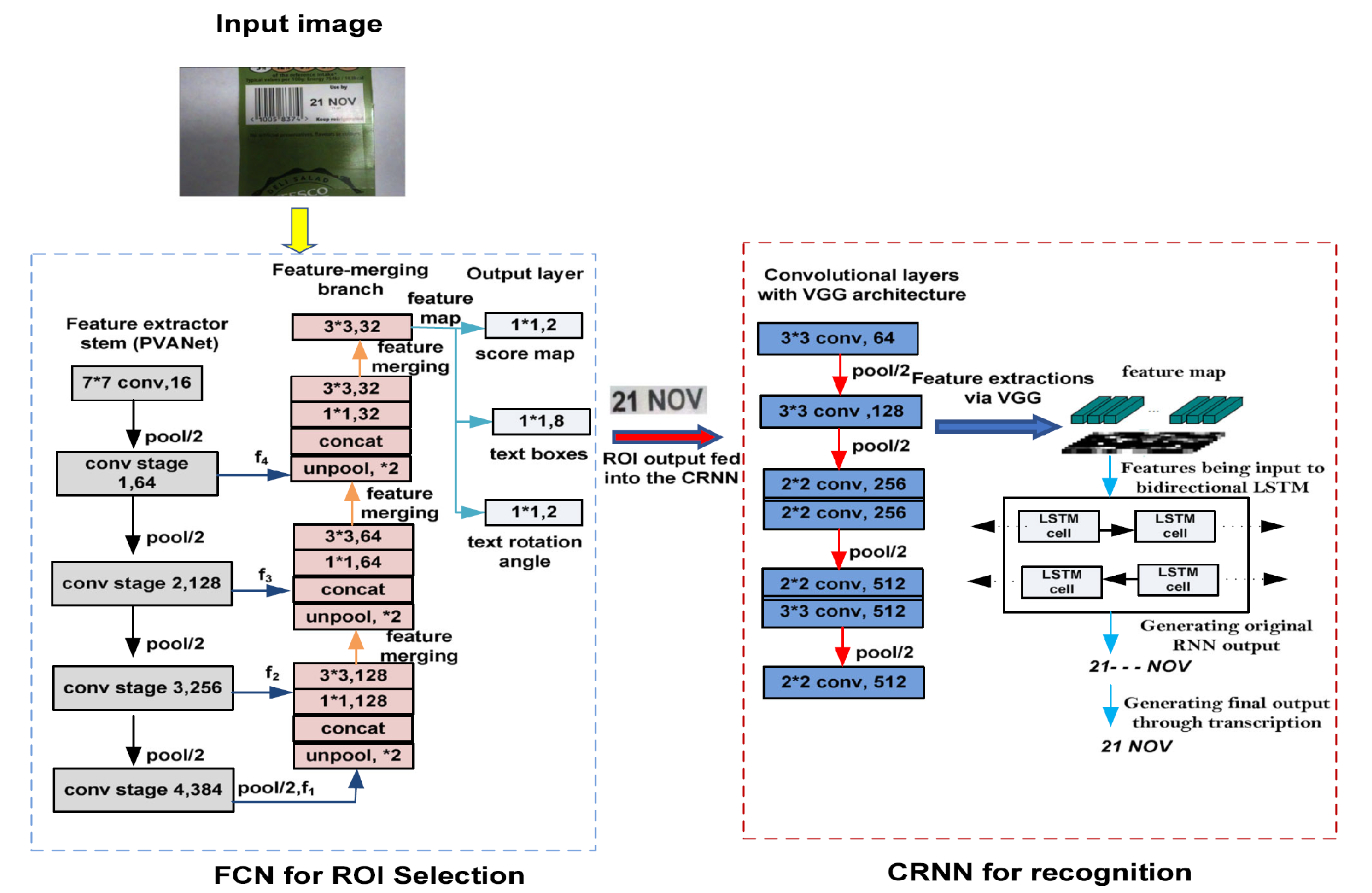}
\centering
\caption{Architecture of the FCN-CRNN system}
\label{fcn-rnn}
\end{figure*}

The FCN-RCNN architecture is shown in Figure \ref{fcn-rnn}. 
It includes the FCN part,
which is responsible for the detection of the Region-Of-Interest (ROI) of the expiry
date. This acts as a filter to identify the image patch including the ROI
from a whole food package photograph, so that
the recognition task is performed on that
specific small image patch, instead of the whole image.
The second part includes the CRNN network, which
performs date character recognition on the image
patch obtained from the first network. Multiple levels of features
are extracted from equally divided regions in the expiry date ROI, so as to recognize the characters within each region,
while contextual relationships between characters in consecutive
regions are modelled by the
recurrent layers of CRNN.

The accuracy of expiry date  detection by the fine-tuned FCN model reaches 98,2 \%. It is higher than the accuracy obtained with two other similarly fine-tuned popular deep neural networks, i.e., CTPN \cite{ref36} and Seglink \cite{ref37}, while providing much fewer false alarms and miss detections, as shown in Table 2. 

\begin{table}[tph!]
\caption{Performance of the FCN detection sub-system }
\label{table:fcn}
\centering
\begin{tabular}{|c|c|c|c|}
\hline 
Method & Missing Detection (\%) & False Alarm (\%) & Accuracy (\%) \\
\hline \hline
FCN & 1.67 & 0.28 & 98.20\\
CTPN \cite{ref36} & 2.79 & 16.57 & 92.20\\
Seglink \cite{ref37} & 5.71 & 12.53 & 93.73\\
\hline 
\end{tabular}
\end{table}

Then the CRNN part of the method was fine-tuned using annotated (rectangular) image patches extracted by the FCN part.  As shown in Table 3, the achieved recognition accuracy is 95,44 \%, much higher than that provided by the Tesseract OCR tool \cite{ref39} and higher than the performance of the TPS-ResNet-BiLSTM-Att network \cite{ref38}, which is much more complex (containing 6 times more parameters than the CRNN model).

\begin{table}[tph!]
\caption{Performance of the CRNN recognition sub-system }
\label{table:crnn}
\centering
\begin{tabular}{|c|c|}
\hline 
Method & Accuracy (\%) \\
\hline \hline
CRNN  & 95.44\\
TPS-ReSNet-BiLSTM-Att \cite{ref38} & 94.57\\
Tesseract OCR \cite{ref39} &  31.12\\
\hline 
\end{tabular}
\end{table}

\subsubsection{Latent Variable based Expiry Date Verification}

A high variability exists in the image characteristics captured in different environments. As a consequence, an FCN model trained with images from a dataset collected in one location does not perform well when applied to images collected in another location. A method to cope with this problem is by extracting latent variables from each trained FCN; clustering them as described in Section 2.6.2 to produce respective cluster centroids; using these centroids as an equivalent model for evaluating images from other locations through nearest neighbor classification \cite{ref40}. 

It has been shown that best results are achieved in this way, if latent variables are extracted by a dense layer preceding the output network layer \cite{ref2, ref41, ref42}. For this reason, some dense layers are included on top of the FCN network, as shown in Figure 10. In particular, an averaging pooling and two dense layers, each containing 2048 ReLU units are added and the 2048 neuron outputs of the last dense layer are clustered using the k-means algorithm. 

Let us consider the case of two datasets obtained from different locations. 7 clusters are generated per dataset cite{ref40}, after applying the above procedure to each dataset. The respective 14 centroids are then merged, following an algorithm based on  \cite{ref2, ref50}, which Iteratively removes the lowest performing cluster, resulting in adaptation of some of the remaining centroids, until the expiry date verification stops improving.

Figure 11 shows the effect of the algorithm on the 14 original cluster centoids. Two of them (black stars) were excluded; four centroids (red squares) in class 1 and two (green circles) in class 2 were adapted and the rest CNN centroids remained unaffected. It should be mentioned that this produced an improvement in the verification accuracy of 13.8 \% in total accuracy over the datasets of the two locations.

\begin{figure*}[tph!]
\includegraphics[totalheight=3cm]{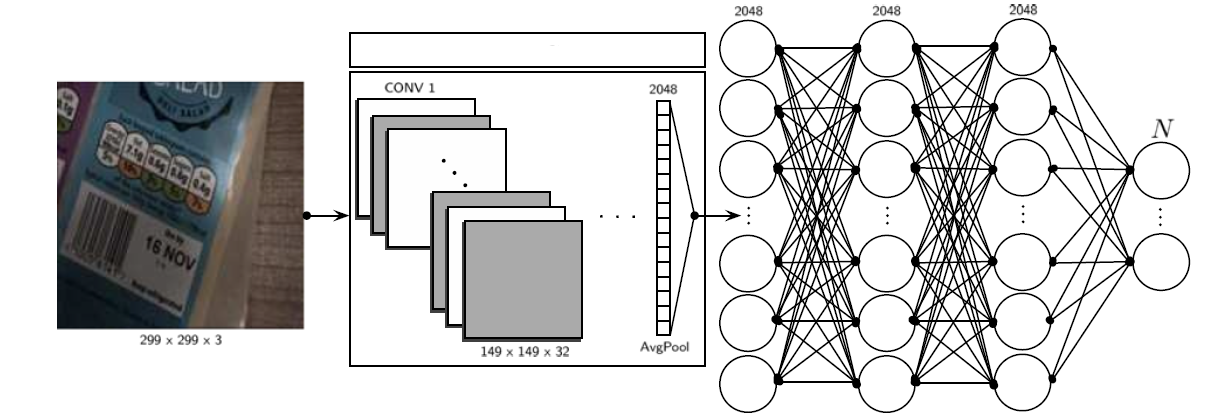}
\centering
\caption{FCN part followed by dense layers - clustering the last layer neurons outputs}
\label{dataset}
\end{figure*}

\begin{figure*}[tph!]
\includegraphics[totalheight=5cm]{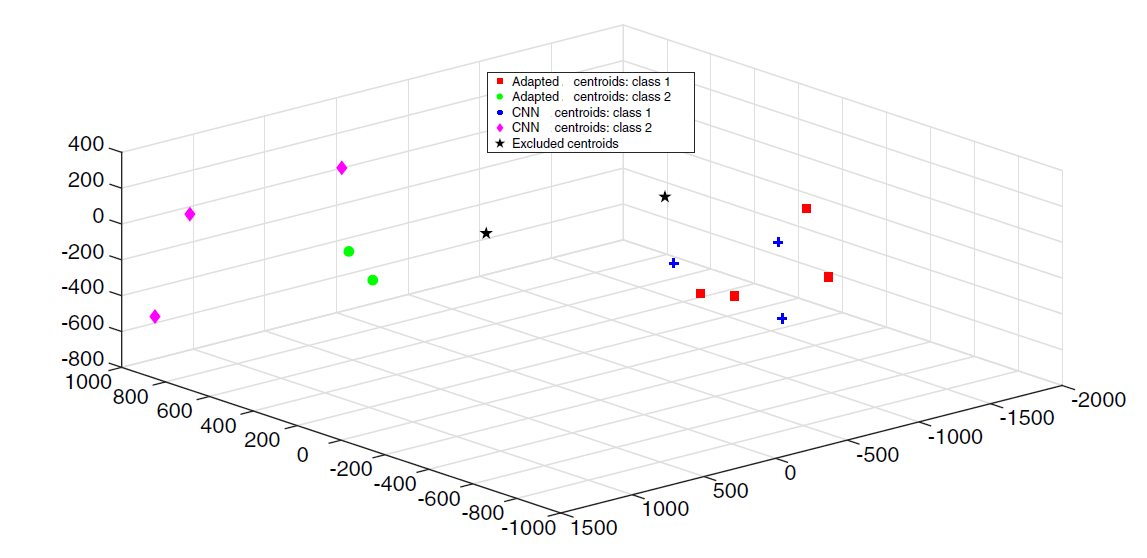}
\centering
\caption{3-D Visualization of the adapted, constant and excluded cluster centroids; class 1(2) includes good(bad) quality images}
\label{centroids}
\end{figure*}

\subsubsection{Domain Adaptation for Multi-source Expiry Date Recognition}

The problem tackled in the former subsection, of high variability across datasets collected in different environments, as well as the general unavailability of labeled data across the whole food supply chain, is further examined next, using the domain adaptation approach described in Section 2.7.

Figure 12 shows the developed domain adaptation procedure.  It comprises a feature extractor and
a classification part. The feature extractor part learns useful
representations, with its sub-networks
learning features specific to each source-target domain pairs.
The classification part of the model learns domain-specific
attributes for each target image. A Class Activation Map is included, as described in Section 2.6.1 to provide 
visualization of which parts of the input images affected most the provided predictions.

\begin{figure*}[tph!]
\includegraphics[totalheight=8.5cm]{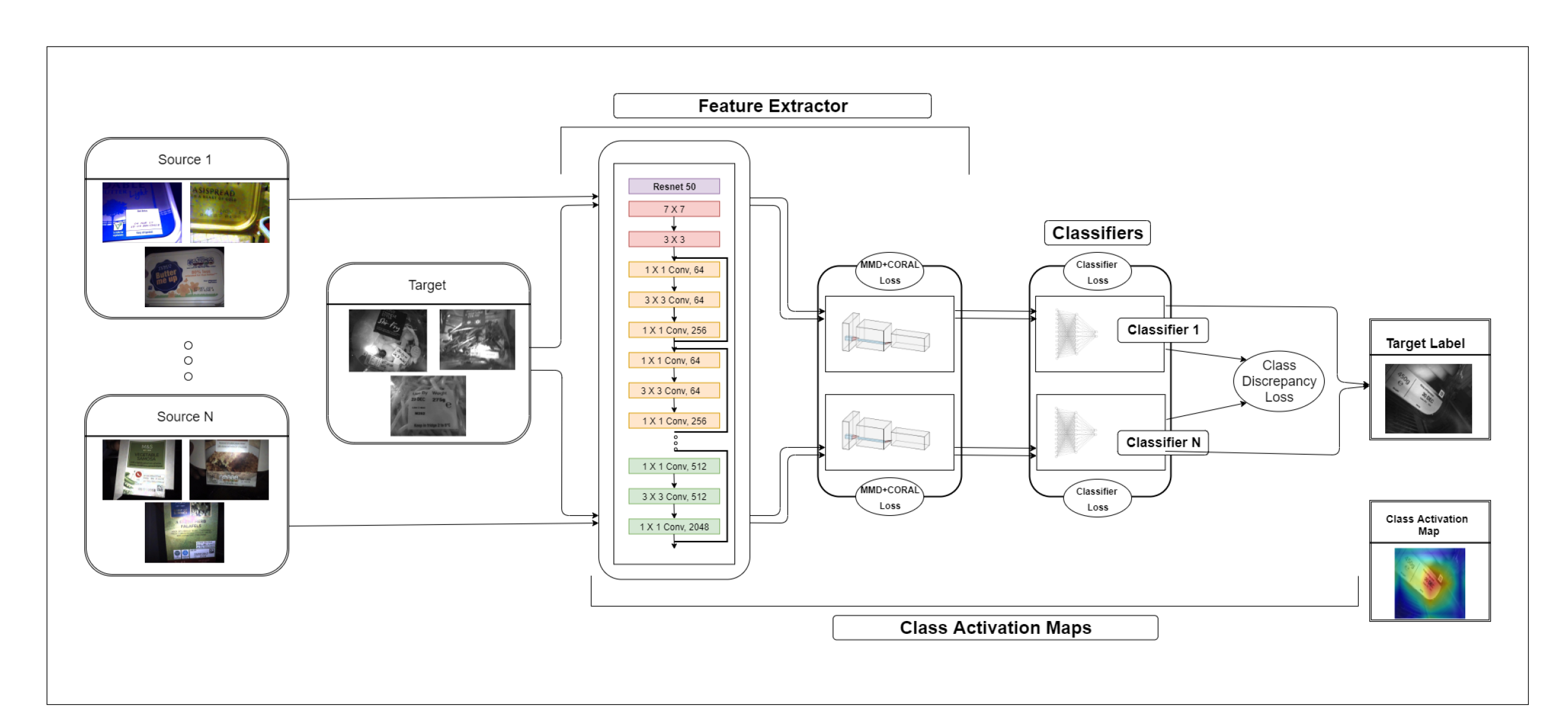}
\centering
\caption{The Multi-source Domain Adaptation architecture}
\label{Dafig}
\end{figure*}

Experiments were made using a labeled single source
dataset and an unlabeled single target dataset for all six locations. The goal of this experiment has been to establish
a baseline for images that would be classified as readable
and acceptable according to human standards. Further experiments
were conducted using the multi-source
domain adaptation approach, by using either two labeled
source datasets and a single unlabeled target domain, or three
labeled source datasets with a single unlabeled target domain.

For comparison purposes, additional
experiments were carried out by combining the two, or the three
source datasets into a single source dataset and adapting them towards the single target domain dataset. 
In the experiments, either the FCN - or the ResNet-50 \cite{ref43} pretrained on ImageNet \cite{ref44} - was the backbone network, by adding,  a dense layer, and  fine-tuning
all convolutional and pooling layers.

\begin{table}[!t]
\caption{Comparison of Performance of Domain Adaptation (DA) Methods }
\label{table:da}
\centering
\begin{tabular}{|c|c|}
\hline 
Method & Accuracy (\%) \\
\hline \hline
Single Source DA & 84.14\\
Two Source Combined DA  & 85.05\\
Three Source Combined DA &  86.13\\
Multi(2)-Source DA & 90.53\\
Multi(3)-Source DA & 92.50\\
\hline 
\end{tabular}
\end{table}

Table 4 presents the average classification accuracy for all examined domain adaptation methods. 
As shown in the Table, the results of source combined methods are
comparatively better than single source methods. This can be justified by considering this case as data enrichment, indicating
that combining multiple source domains into single source
domain is helpful. The multi-source
domain adaptation approach significantly outperforms both other methods with an average classification accuracy improvement by
more than 6 \%.







\section{Discussion}

The presented experimental studies illustrated that the use of AI and ML methodologies can provide the food supply chain with efficiency and safety, reducing food waste and environment pollution; this is the main target of the data platforms required for food systems \cite{ref52, ref53}.
Most of the presented work has taken advantage of the recent progress in deep learning and deep neural networks. 

In the first  case study, DL systems were trained to predict growth and yield in time series data collected in greenhouses, taking into account environmental data and historical growth, or yield data. It has been shown that RNN/LSTM models combined with autoencoding, or attention mechanisms, were able to provide state-of-the-art performance compared with existing  methods for multi-step prediction of stem diameter based growth of plants, as well as of tomato yield weekly prediction.  Similar approaches can be used for prediction of yield of other categories, for example fruits, and in particular strawberry yield and their harvesting using robots \cite{ref54, ref55}. 

In the second case study, it has been shown that control of the energy consumed  by big food suppliers, e.g., large supermarket companies, is feasible, using deep learning based prediction. A large scale  study, for automatic control, i.e., temporary shutdown of retailing refrigerator systems in peak energy demand hours, has been implemented, based on prediction of the specific refrigerator defrost period in each considered location. RNN/LSTM models were mainly used to perform the prediction task; GAN models were also successfully used to perform data augmentation and prediction. Similar methods have been used to predict  anomalies, i.e., instances  when  the  power demand  load  gets  higher  than  the available power in a country \cite{ref61}. 

Current targets related to food supply chain, worldwide, include \cite{ref62}: cut of greenhouse gas emissions by at least 55 \% by 2030; development and use of energy-efficient approaches; optimization of connectivity to energy; analysis of production data; efficient management of resources, such as energy, water, soil,  biodiversity. The presented approaches illustrate how AI and DL can be used to assist towards these goals.

The third case study illustrated that automatic recognition of expiry date in retail food packaging
can also be successfully implemented through deep learning methodologies. The impact of this task is also crucial for ensuring food safety and consumers' health, whilst reducing food waste and related financial loss. 

It was shown that FCN and CRNN models can be used to effectively recognize the expiry date on real life food packaging. Verification of the expiry date, with visualization of the decision making procedure can be achieved through latent variable extraction of the trained deep neural architectures and a related clustering procedure. Moreover, domain adaptation of the deep neural architectures can be performed so as to achieve expiry date verification and recognition across different retail environments.  

The presented deep learning approaches can
also be applied on wider aspects of food package control,
such as the verification of the allergen labeling barcode, or other
nutritional information. These can create a further impact towards people's safety, by automatically informing customers which allergens are included in the food products they are thinking of buying in retail supermarkets.   

All above experiments have used the state-of-the-art in deep learning methodologies. In future research, the developments will focus on extending the described frameworks, with a target to  combine deep learning with AI knowledge representation and reasoning, as well as multi-objective optimization technologies \cite{ref63, ref64}.  

In this context, the methodologies that were presented in this paper for achieving an efficient and safe food supply chain will be extended, so as to create trustworthy AI-enabled food chain supply, that offer transparency and explainability to their users, i.e., food producers and farmers, retail supermarket owners, customers and the general public.

In particular, the deep neural architectures, especially the approach deriving the backward model of extracted cluster characteristics, described in Section 3.3.2, can be interweaved with ontological representation \cite{ref97} of this backward model, so as to provide explainable cues of the decision making procedure by using query answering techniques \cite{ref65, ref66}. 

Moreover, the derived models will be  adapted and contextualized across different environments, using general-purpose methodologies, as developed for other classification and prediction fields \cite{ref67, ref68}.
In addition, simultaneous maximization of yield and minimization of the consumed energy in greenhouses can be tackled through dynamic multi-objective optimization \cite{ref96, ref69}.


\vspace{6pt} 



\authorcontributions{The individual contributions of the authors are as follows: Conceptualization, Stefanos Kollias; methodology, Ilianna Kollia; software, Jack Stevenson; validation, Stefanos Kollias, Jack Stevenson; formal analysis, Ilianna Kollia; writing---original draft preparation, Ilianna Kollia and Jack Stevenson; writing---review and editing, Stefanos Kollias. All authors have read and agreed to the published version of the manuscript.}





\appendixtitles{no} 


\end{paracol}

\reftitle{References}
\externalbibliography{yes}
\bibliography{bibtex}

\end{document}